%% file: root.tex
\newcommand{\CLASSINPUTtoptextmargin}{19.1mm}
\newcommand{\CLASSINPUTbottomtextmargin}{19.1mm}
\newcommand{\CLASSINPUTinnersidemargin}{19.1mm}
\newcommand{\CLASSINPUToutersidemargin}{19.1mm}
\documentclass[conference]{IEEEtran}
\IEEEoverridecommandlockouts
\usepackage{cite}
\usepackage{amsmath,amssymb,amsfonts}
\usepackage{graphicx}
\usepackage{textcomp}
\usepackage{xcolor}
\usepackage{svg}
\usepackage{url}
\def\BibTeX{{\rm B\kern-.05em{\sc i\kern-.025em b}\kern-.08em
    T\kern-.1667em\lower.7ex\hbox{E}\kern-.125emX}}

\newcommand{\supplementarytitle}{%
    \section*{\centering \LARGE \bfseries Supplementary Material}%
    \addcontentsline{toc}{section}{Supplementary Material} 
}

\newcommand{\spatialfun}[2]{\gamma_{#1}(#2)}

\usepackage{algorithm}
\usepackage{algpseudocode}

\usepackage{fancyhdr}
\begin{document}
\pagenumbering{arabic}

\title{\vspace{8mm}Lab2Car: A Versatile Wrapper for Deploying Experimental Planners in Complex Real-world Environments
\thanks{$\dagger$ -- work done while at Motional. }}

\author{\IEEEauthorblockN{
  Marc Heim,
  Francisco Su\'{a}rez-Ruiz$^\dagger$,
  Ishraq Bhuiyan, 
  Bruno Brito$^\dagger$,
  Momchil S. Tomov
  }
   \IEEEauthorblockA{\textit{Motional AD Inc.}\\
  \texttt{\{marc.heim,ishraq.bhuiyan,momchil.tomov\}@motional.com}}
}

\maketitle

\thispagestyle{fancy}

\begin{abstract}

Human-level autonomous driving is an ever-elusive goal, with planning and decision making -- the cognitive functions that determine driving behavior -- posing the greatest challenge. Despite a proliferation of promising approaches, 
progress is stifled by the 
difficulty of 
deploying experimental 
planners in naturalistic settings. In this work, we propose Lab2Car, an optimization-based wrapper that can take a trajectory sketch from an arbitrary motion planner and convert it to a safe, comfortable, dynamically feasible trajectory that the car can follow. 
This allows motion planners that do not provide such guarantees to be safely tested and optimized in real-world environments.
We demonstrate the versatility of Lab2Car by using it to deploy a machine learning (ML) planner and a classical planner on self-driving cars in Las Vegas. The resulting systems handle challenging scenarios, such as cut-ins, overtaking, and yielding, in complex urban environments like casino pick-up/drop-off areas. Our work paves the way for quickly deploying and evaluating candidate motion planners in realistic settings, ensuring rapid iteration and accelerating progress towards human-level autonomy.
\end{abstract}


\section{Introduction}

Self-driving cars have achieved remarkable progress towards human-level autonomous driving. 
Much of this success is owed to progress in ML-based perception and prediction \cite{Zhou_2018_CVPR, Lang_2019_CVPR, vora_2020_cvpr, phan-minh_2020_cvpr, yin_2021_cvpr, chen_2021_neurips, liang_2020_eccv}, which can attain a human-like understanding of the scene around the autonomous vehicle (AV). 
Classical motion planners relying on handcrafted rules \cite{lavalle2006planning,Paden_2016} have similarly given way to ML-based motion planners that learn the rules of driving from data \cite{scheel2022urban,phan2023driveirl,Bojarski_2016_arxiv, Codevilla_2019_ICCV, Bansal_2019_RSS, Zeng_2019_CVPR, Hawke_2020_ICRA, Vitelli_2022_ICRA, Hu_2023_CVPR}. ML planning is thought to be more scalable and better positioned to capture the ineffable nuances of human driving behavior than classical planning. 

However, deploying ML planners in the real world comes with its own set of challenges, which are often overcome using techniques from classical planning. For one, na\"ive trajectory regression does not ensure comfort or even basic kinematic feasibility, necessitating post-hoc smoothing \cite{chen2019,naveed2021,wabersich2021} or hand-engineered trajectory generation \cite{phan2023driveirl,Vitelli_2022_ICRA}. 
Ensuring safety poses an even greater challenge, as ML solutions tend to fail on edge cases that compose the long tail of the data distribution. Since the opacity of the implicitly learned rules makes it difficult to predict when such failures would occur, it is of paramount importance to institute interpretable guardrails when deploying ML planners in the real world. Previous authors have achieved this by projecting the ML trajectory onto a restricted set of lane-follow trajectories pre-filtered based on hand-engineered rules \cite{phan2023driveirl,Vitelli_2022_ICRA}, an ad-hoc approach which does not scale to complex, unstructured scenarios. Other authors \cite{karkus2023,huang2023} have proposed applying differentiable model predictive control (MPC) as the final layer of a neural AV stack to simultaneously ensure safety, comfort, and feasibility. 
However, these approaches were not deployed in the real world and come at the cost of increased complexity and tight coupling between the MPC and the planning module. 

In all instances, making ML planners or even na\"ive classical planners deployable requires substantial engineering investment, which can only be justified if there is strong signal that they will drive well. Such signal could in principle be obtained in simulation; however, there is often a substantial gap between performance in simulation and on the road. As a result, few experimental planners get evaluated in the real world, and those that do are often part of bespoke deployment systems that cannot be readily extended to other planners.

\begin{figure}
    \centering
    \includegraphics[width=0.49\textwidth,trim={0 0 0 40},clip]{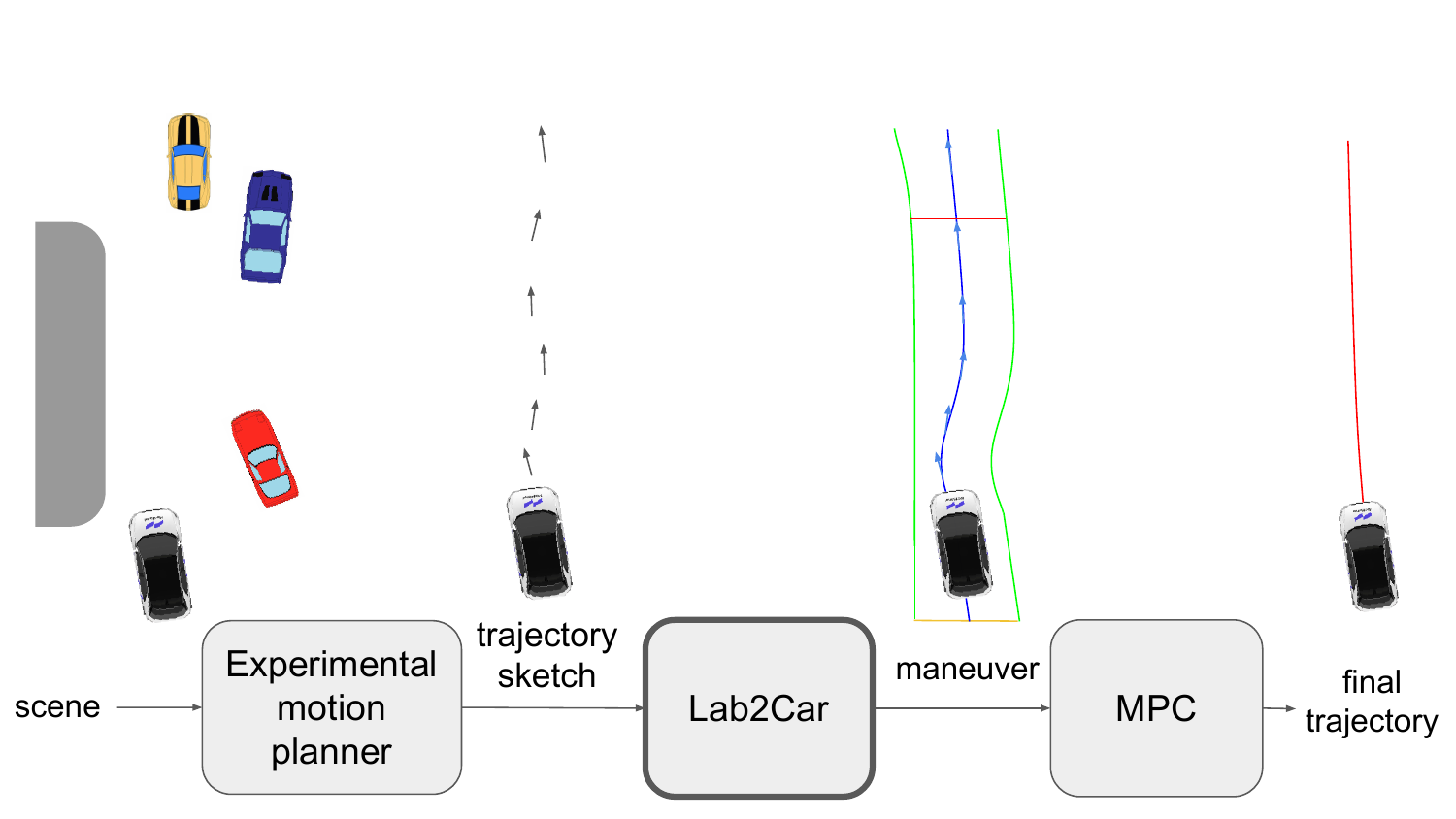}
    \caption{Lab2Car is a plug-and-play wrapper that transforms a rough trajectory sketch 
    into spatiotemporal constraints (a maneuver). The maneuver defines an optimization problem solved by MPC to obtain a final trajectory that is safe, comfortable, and dynamically feasible. This enables rapid deployment and real-world evaluation of 
    planners that lack such properties.}
    \label{fig:arch}
\vspace{-5mm}
\end{figure}

To address this problem, we present Lab2Car, an optimization-based wrapper that takes a rough trajectory sketch 
and transforms it into a set of interpretable spatiotemporal constraints -- a \textit{maneuver} -- which is then solved using MPC to obtain a safe, comfortable, and dynamically feasible trajectory (Fig.~\ref{fig:arch}). 
The initial trajectory sketch captures the high-level behavioral 
intent, 
while the final 
trajectory captures the low-level sequence of commands that the robot must execute. 
This division of labor mirrors hierarchical motor control in biological brains \cite{merel2019hierarchical} and allows for robust and safe deployment of a wide range of experimental planners. 


Our contributions are as follows:



\begin{itemize}
    \item A simple yet powerful maneuver extraction algorithm that converts a trajectory sketch -- even a really poor one -- into constraints that preserve the underlying driving intent while ensuring safety.
    \item Integration with multiple ML/classical planners and a state-of-the-art MPC solver as part of a full AV stack.
    \item Evaluation on 2000+ resimulated scenarios from real-world drive logs using a high-fidelity simulator.
    \item Deployment on a real AV and evaluation on the road.
\end{itemize}




\begin{figure}
    \centering
    \includegraphics[width=0.5\textwidth,trim={0 130 25 0},clip]{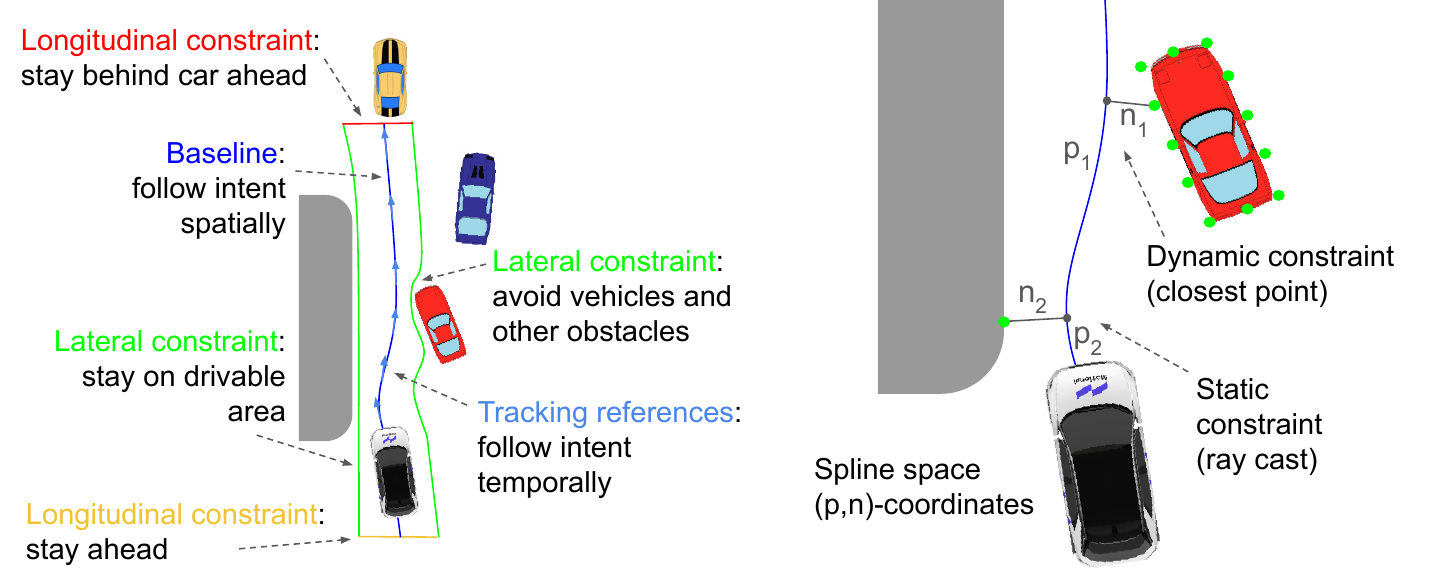}
    \caption{Anatomy of a maneuver (left) and spline-space illustration (right).}
    \label{fig:maneuver}
\end{figure}

\begin{figure}
    \centering
    \includegraphics[width=0.5\textwidth,trim={20 40 280 0},clip]{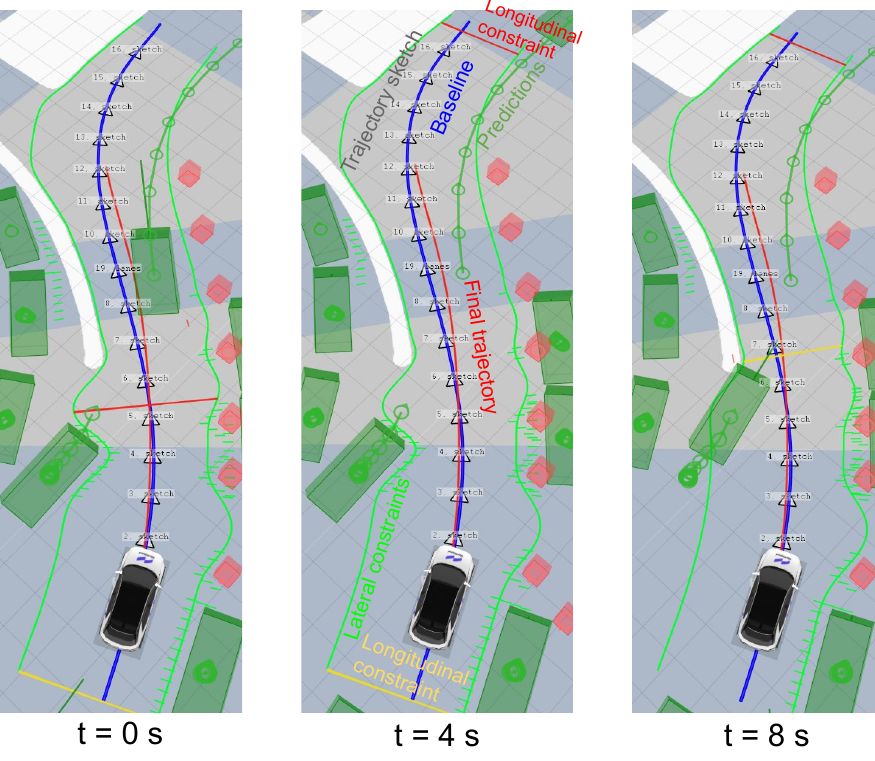}
    \caption{Lab2Car in real-world scenario. Gray arrowheads denote initial trajectory sketch from experimental planner. Spatiotemporal constraints denoted as in Fig.~\ref{fig:maneuver}. Connected dark green circles denote predictions for other agents. Red curve denotes final 8-s open-loop trajectory from MPC.}
    \label{fig:full1}
\vspace{-5mm}
\end{figure}

\section{Maneuver definition}

The input to Lab2Car is a path or trajectory sketch $\tau = \{(x_k, y_k, t_k)\}_{k=0}^{K-1}$ consisting of a sequence of $K$ Cartesian waypoints $(x_k, y_k)$ with optional timestamps $t_k$. The trajectory is then converted to a maneuver: a set of spatiotemporal constraints (continuous in the spatial domain and discrete in the temporal domain) that define the feasible space of trajectories over the planning horizon (Fig.~\ref{fig:maneuver}, left).  
Similarly to control barrier functions \cite{ames2019control,notomista2020,seo2022}, the maneuver implicitly defines a safe region, while also optimizing for comfort and similarity to the trajectory sketch.

\textbf{The baseline} trajectory is modeled as a 2D quartic cardinal B-spline \cite{piegl2012nurbs} (Fig.~\ref{fig:maneuver} and Fig.~\ref{fig:full1}, blue curve). The spline is defined by a set of $N$ control points and corresponding knot vectors. The knot vectors are placed at spline progress values $p$ from \(0\) to \(N\), 
The spline basis functions are valid within the domain spanning from \(1.5\) to \(N-2.5\), ensuring that the spline is smooth and well-defined across this range. 


This baseline spline serves as the reference axis of a curvilinear coordinate system which defines all subsequent trajectory constraints in terms of longitudinal (along the spline) and lateral (orthogonal to the spline) components.

\textbf{Tracking references} 
are defined as a time series $(p_\text{ref}(t), v_\text{ref}(t), a_\text{ref}(t))$, specifying the progress, longitudinal velocity, and longitudinal acceleration, respectively, that the AV should aim to achieve at each time point $t$ of the discretized planning horizon (Fig.~\ref{fig:maneuver}, blue arrows). These do not apply if $\tau$ is a path (i.e., has no timestamps).

\textbf{Lateral constraints} are formulated as 1D spline functions extending along the baseline spline, representing the driveable space boundaries at each time point $t$ (Fig.~\ref{fig:maneuver} and Fig.~\ref{fig:full1}, green curves). These constraints are segmented into ``hard'' and ``soft'' constraints for both the left and right boundaries, 
indexed by spline progress: $
        \spatialfun{\text{left,hard}}{p}, \spatialfun{\text{left,soft}}{p}, \spatialfun{\text{right,hard}}{p}, \spatialfun{\text{right,soft}}{p} $.
A separate set of 
constraints is defined for each time point $t$ (omitted). 

\textbf{Longitudinal constraints}  
are defined as a time series $(p_{\text{lower}}(t), p_{\text{upper}}(t))$ specifying the lower and upper bound on progress, respectively, at each time point $t$ of the planning horizon (Fig.~\ref{fig:maneuver} and Fig.~\ref{fig:full1}, yellow and red lines). 


\begin{figure*}
    \centering
    \includegraphics[width=1\textwidth,trim={0 248 170 0},clip]{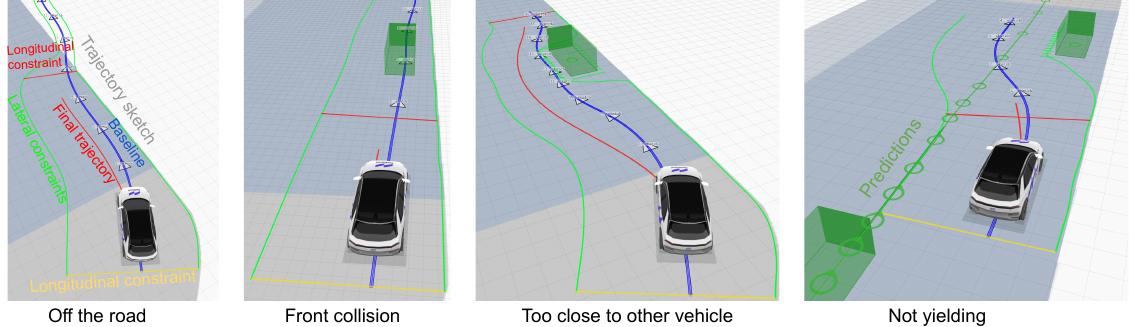}
    \caption{Lab2Car rescuing bad trajectory sketches in synthetic scenarios. Color coding as in Fig.~\ref{fig:full1}}
    \label{fig:examples}
\vspace{-5mm}
\end{figure*}

\section{Maneuver extraction}

\subsection{Baseline fitting}

Fitting the baseline 
proceeds in two steps.

\textbf{Step 1: Determining spline progress}:
Initially, the progress value \(\hat{p}_k\) for each waypoint point along the spline is calculated to reflect the cumulative distance along the path:

\begin{equation}
    \hat{p}_k = \begin{cases} 
    0 & \text{if } k = 0\\ 
    \hat{p}_{k-1} + \frac{\|\mathbf{w}_{k-1} - \mathbf{w}_k\|}{\text{dist}_\text{max}} & \text{otherwise}
    \end{cases}
\end{equation}

Here, $\mathbf{w}_k = \begin{bmatrix}  x_k, y_k  \end{bmatrix}$ are waypoint vectors and \(\text{dist}_\text{max}\) is the target 
distance between control points on the spline.

\textbf{Step 2: Control point optimization}:
The control points $C$ of the spline are computed using a least squares method by solving: 

\vspace{-7mm}

\begin{equation}
\begin{bmatrix}
\mathbf{w}_0, \mathbf{w}_1, \dots, \mathbf{w}_{K-1}, 0, 0, \dots, 0
\end{bmatrix}^T
\approx
\begin{bmatrix} B \\ \hline \\ c_{reg} R \end{bmatrix} \cdot C 
\label{eq:control_point_optimization}
\end{equation}

\begin{equation}
B_{K \times N} = \begin{bmatrix}
\mathbf{b}(\hat{p}_0), \mathbf{b}(\hat{p}_1), \dots, \mathbf{b}(\hat{p}_{K-1})  
\end{bmatrix}^T
,
\end{equation}


where \( C \) is a $N \times 2$ matrix of control points, $\mathbf{b}$ is the spline basis function, $R$ is a second-order regularization matrix which penalizes excessive curvature, and \( c_{reg} \) is the regularization coefficient. 

\subsection{Projection of dynamic obstacles onto baseline}

Detections and predictions for other dynamic road entities, such as vehicles and pedestrians, are represented by time series of convex hulls subsampled to sets of $(x, y)$ points.
The sampled points are then transformed into a curvilinear coordinate system aligned with the baseline, referred to as ``spline-space'' (Fig.~\ref{fig:maneuver}, right). In this coordinate system, each point is expressed as $(p,n)$, where $p$ denotes progress of the closest point along the spline and $n$ represents the lateral deviation from the spline. 
The resulting $(p,n)$-points determine the \textit{dynamic constraints}.

\subsection{Projection of static obstacles onto baseline}

The map is represented by a 
raster in which each voxel is labeled as ``drivable'' or ``non-drivable''. At uniformly sampled points along the baseline, a ray cast perpendicular to the baseline in both directions finds the nearest non-drivable voxels in the map raster (Fig.~\ref{fig:maneuver}, right). The resulting $(p,n)$-points determine the \textit{static constraints}.

Note that a similar procedure could be used for the dynamic obstacles if they were rasterized rather than vectorized. 

\subsection{Constraint computation}

\textbf{Tracking references} are computed by projecting each waypoint $(x_k, y_k)$ onto the baseline to obtain $p_\text{ref}(t_k)$. Longitudinal velocity $v_\text{ref}(t)$ and acceleration $a_\text{ref}(t)$ references are approximated using finite differences.

\textbf{Dynamic constraints} are determined as follows:

\begin{itemize}
  \item Dynamic obstacles farther than a certain threshold (i.e., whose closest point has $n$ greater than the threshold) impose lateral constraints only.
  \item Dynamic obstacles closer than the threshold can impose both lateral and longitudinal constraints, determined separately for each  point based on another threshold.
\end{itemize}

\textbf{Static constraints} are lateral by default, unless $n$ is below a certain threshold, in which case they are longitudinal.

\textbf{Stay-behind or stay-ahead}:
Longitudinal constraints can be imposed differently depending on the assumptions about the experimental planner. We consider two regimes:

\begin{itemize}
    \item \texttt{Stay-behind}: the \textit{initial} position of the AV must always fall within the longitudinal bounds. Formally, $p_\text{lower}(t) < p_\text{rear}(0) < p_\text{front}(0) < p_\text{upper}(t) \,\,\,\forall t$, where $p_\text{rear}(t), p_\text{front}(t)$ are the progress values of the rear bumper and the front bumper of the AV at time $t$, respectively. In other words, the AV always \textit{stays behind} (i.e. yields for) dynamic obstacles predicted to cross its path. This is suitable for planners that output a path (i.e., no timestamps) or that do not take dynamic road actors into consideration.

    \item \texttt{Stay-ahead}: the \textit{planned} position of the AV must always fall within the longitudinal bounds. Formally, $p_\text{lower}(t) < p_\text{rear}(t) < p_\text{front}(t) < p_\text{upper}(t) \,\,\,\forall t$. In this regime, the AV may \textit{stay behind or stay ahead} of other actors, depending on its predicted relative position along the baseline. This is suitable for planners that output trajectories and take other road actors into consideration.
\end{itemize}

\subsection{Lateral constraint fitting}

The lateral splines are fitted to the samples using quadratic programming  \cite{wright2006numerical}, which  maximizes the available space within the lateral constraints, subject to $\spatialfun{\text{left,hard}}{p_i} < n_i$ and $\spatialfun{\text{right,hard}}{p_i} > n_i$ for every laterally constraining sample point $(p_i, n_i)$. Regularization penalizes excessive curvature, ensuring smoothness of the resulting splines. 

\section{MPC formulation}

The dynamical system is defined by the state $\mathbf{x}(t) = [p(t), n(t), \omega(t), v(t), a(t), \beta(t)]$ and the control inputs $\mathbf{u}(t) = [j(t), \Delta \beta(t)]$, where the components are:

\vspace{2mm}

\begin{tabular}{p{1.5cm} p{6cm}}
\multicolumn{2}{l}{\textbf{State} $\mathbf{x}(t)$} \\ 
$p(t)$ & Progress along the baseline \\ 
$n(t)$ & Lateral error from the baseline \\
$\omega(t)$ & Local heading relative to the baseline \\
$v(t)$ & Longitudinal velocity \\
$a(t)$ & Longitudinal acceleration \\
$\beta(t)$ & Steering angle \\
\multicolumn{2}{l}{\textbf{Control} $\mathbf{u}(t)$} \\
$j(t)$ & Jerk (rate of acceleration change) \\
$\Delta \beta(t)$ & Change in steering angle \\
\end{tabular}

\vspace{2mm}

The MPC problem 
can be formulated as follows:

\begin{equation}
\min_{\mathbf{u}} \sum_{k=0}^{N-1} \ell\left(\mathbf{x}(t+k), \mathbf{u}(t+k)\right) + \ell_f\left(\mathbf{x}(t+N)\right)
\end{equation}

subject to 
$\mathbf{x}(t+1) = f(\mathbf{x}(t), \mathbf{u}(t))$,  
$\mathbf{C}(\mathbf{x}) \leq \mathbf{0}$,
$\mathbf{u}_{\text{min}} \leq \mathbf{u}(t) \leq \mathbf{u}_{\text{max}}  $, and
$\mathbf{x}(0) = \mathbf{x}_0$,


where:
\begin{itemize}
    \item \(\ell(\mathbf{x}(t+k), \mathbf{u}(t+k))\) is the non-linear stage cost function,
    \item \(\ell_f(\mathbf{x}(t+N))\) is the non-linear terminal cost function,
    \item \(N\) is the prediction horizon,
    \item \(f(\mathbf{x}(t), \mathbf{u}(t))\) represents the state transition model -- in this case, a kinematic bicycle model,
    \item \(\mathbf{C}(\mathbf{x})\) represents the non-linear state constraints,
    \item \(\mathbf{u}_{\text{min}}, \mathbf{u}_{\text{max}}\) are the control constraints,
    \item \(\mathbf{x}_0\) is the initial state.
\end{itemize}

The non-linear constraint function \(\mathbf{C}(\mathbf{x})\) includes terms that ensure the 
AV remains within the spline constraints throughout the planning horizon. It also includes 
terms 
balancing various objectives, such as maintaining comfort, ensuring the AV stays within its operational boundaries, and minimizing perceived risks.

\begin{figure}
    \centering
    \includegraphics[width=0.48\textwidth,trim={90 60 100 100},clip]{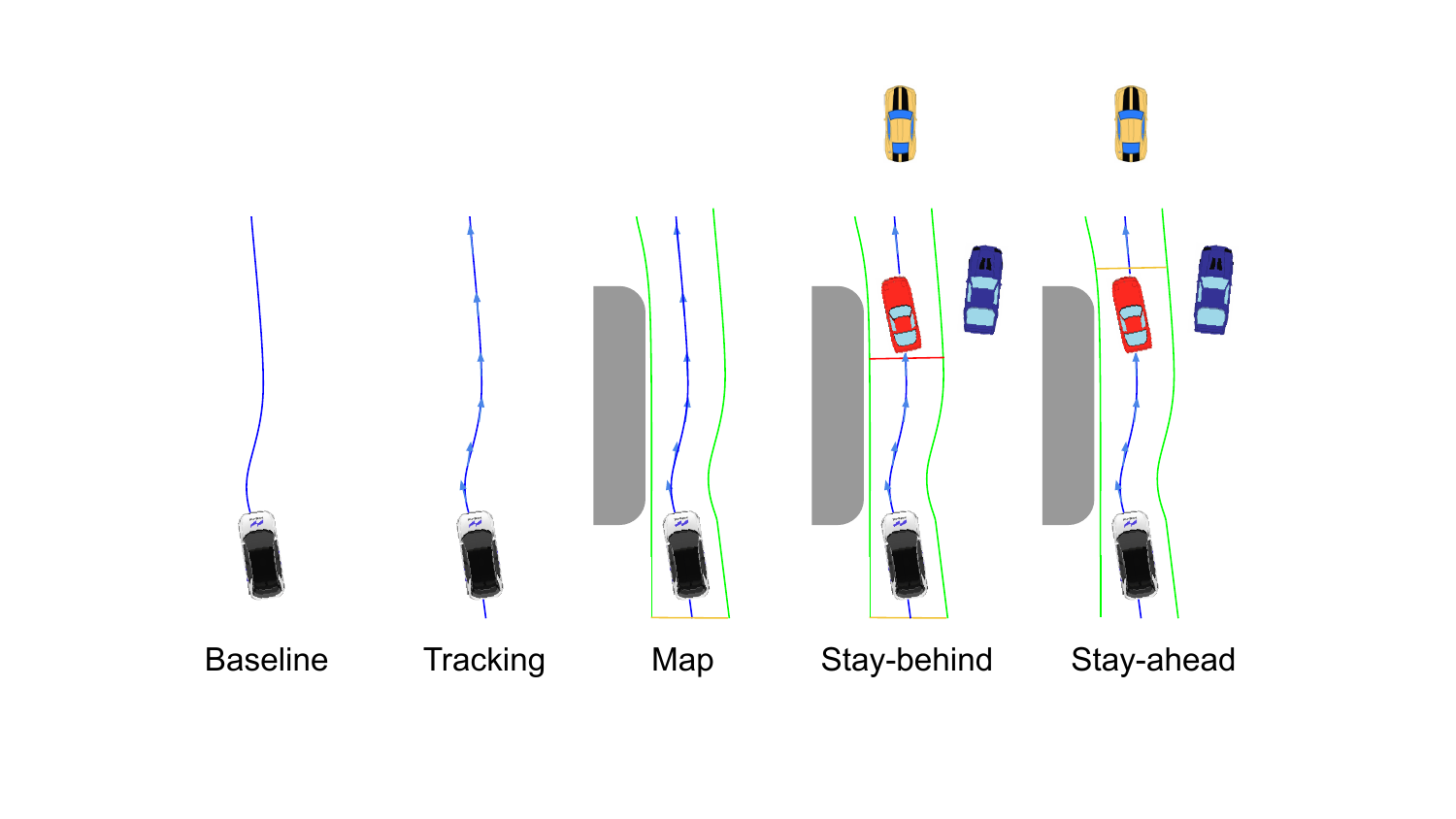}
    \caption{Lab2Car configurations (ablations).}
    \label{fig:ablations}
\end{figure}

\begin{figure*}
    \centering
    \includegraphics[width=1\textwidth,trim={4 100 20 0},clip]{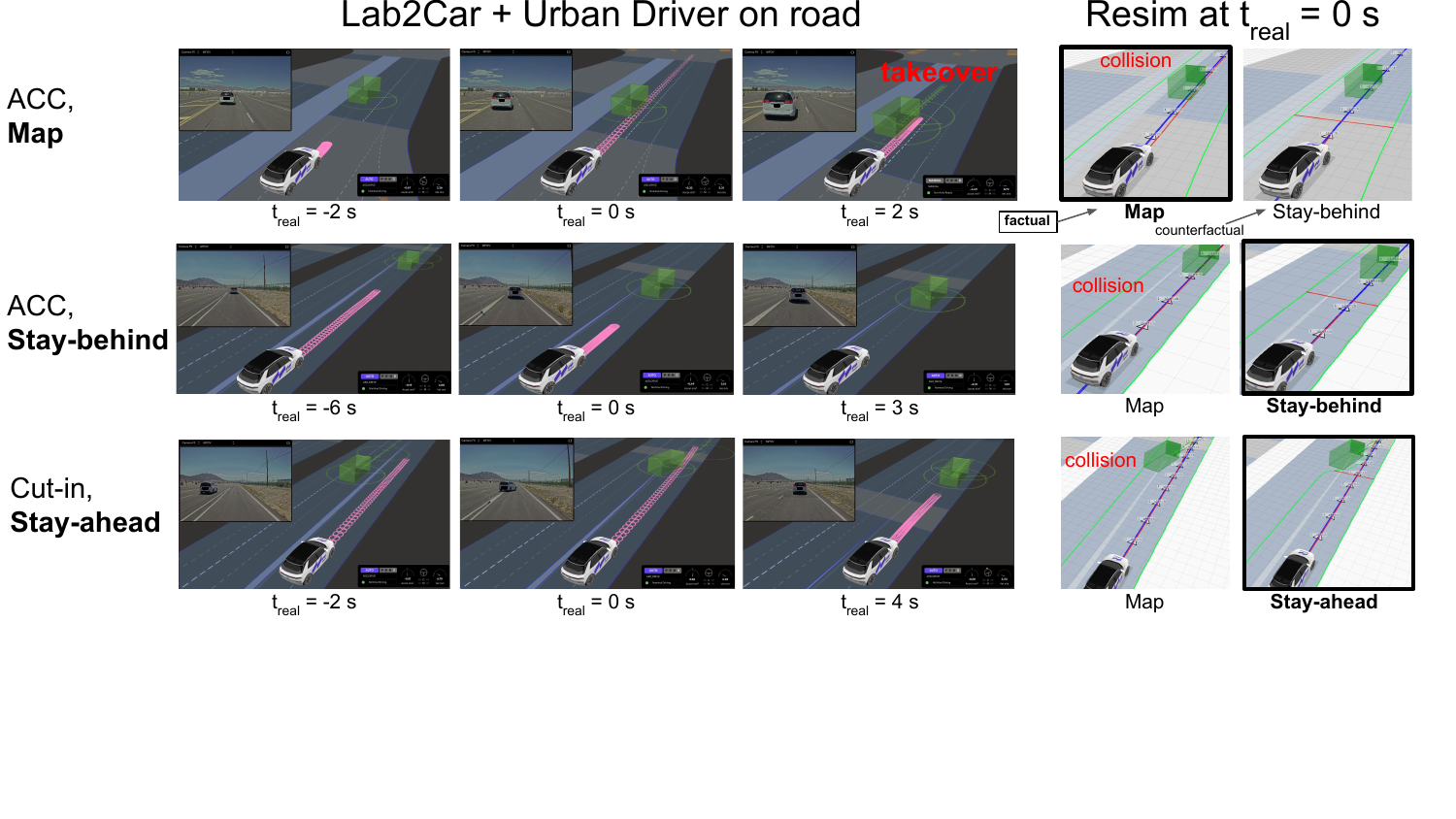}
    \caption{Lab2Car + Urban Driver on the road: sequential 
    frames from example scenarios (rows) in different Lab2Car \textbf{configurations} (left) and corresponding open-loop 
    resims (right) at the reference frame ($t_\text{real} = 0$ s). Each row is from a unique scenario run. 
    Resims performed with \textbf{factual} (bold) configuration (i.e., the one used on the road) and one counterfactual configuration. 
    In left (on-road) panels, open-loop trajectory denoted in pink. In right (resim) panels, color coding as in Fig.~\ref{fig:full1}. Resim, resimulation. 
    See supplementary video for corresponding clips. 
    }
    \label{fig:urbandriveronroad}
\vspace{-3mm}
\end{figure*}

\section{Experiments}
\label{sec:results}

\subsection{Experimental motion planners}



\textbf{Urban Driver} is a regression-based ML planner which learns to imitate human trajectories from expert demonstrations. We trained an open-source version of Urban Driver \cite{caesar2021nuplan,karnchanachari2024towards} 
on lane follow and ACC scenarios from the nuPlan dataset \cite{karnchanachari2024towards}. 

\textbf{Proximal policy optimization (PPO)} is a reinforcement learning method for training an agent to maximize a reward function based on simulated experience. We trained an open-source PPO implementation \cite{stable-baselines3} on lane follow and ACC scenarios using a custom simulator.

\textbf{Intelligent driver model (IDM)} is a classical car-following model which computes longitudinal acceleration in closed form \cite{treiber2000congested}. 

\textbf{A* search} is a classical search-based planner \cite{hart1968formal,lavalle2006planning} which finds an unobstructed path to a goal pose, avoiding stationary vehicles and obstacles. It respects road boundaries but ignores lanes, making it suitable for unstructured environments like casino pick-up/drop-off (PUDO) areas. 

Urban Driver, PPO, and IDM take both static and dynamic obstacles into account and output trajectories, making them suitable for deployment either in the \texttt{Stay-behind} or \texttt{Stay-ahead} configuration. In contrast, A* ignores dynamic obstacles and outputs a path, instead relying on Lab2Car to \texttt{stay-behind} other road actors.

\subsection{Ablations}

In addition to \texttt{Stay-behind} and \texttt{Stay-ahead}, we compared several control Lab2Car configurations (Fig.~\ref{fig:ablations}):

\begin{itemize}
    \item \texttt{Baseline}: Follow the baseline spline, ignoring waypoints and static/dynamic obstacles.
    \item \texttt{Tracking}: Track the waypoints, ignoring static/dynamic obstacles. Only applicable if $\tau$ is a trajectory with timestamps.
    \item \texttt{Map}: Track the waypoints (if applicable) and avoid static obstacles, ignoring dynamic obstacles.
\end{itemize}

\subsection{Simulation results}

\vspace{-3mm}

\begin{table}[htbp]
\centering
\caption{Closed-loop simulation results}
\label{tab:closed_loop_resim}
\begin{tabular}{l|cccc}  
\hline
  &   \multicolumn{2}{c}{\textbf{Safety}} & \multicolumn{1}{c}{\textbf{Comfort}}  & \multicolumn{1}{c}{\textbf{Progress}}\\
\textbf{Config} & Coll (\#)  $\downarrow$ & Road (\#) $\downarrow$ & Accel (\#) $\downarrow$ & Dist (m) $\uparrow$ \\

\hline
\multicolumn{5}{c}{\textbf{Lab2Car + Urban Driver} (lane follow)} \\
\hline
Baseline & 0.551 & 0.021 & 2.030 & \textbf{262.784} \\ 
Tracking & 0.057 & 0.002 & \textbf{0.203} & 214.747 \\ 
Map & 0.063 & \textbf{0.001} & 0.216 & 214.204 \\ 
Stay-behind & 0.010 & \textbf{0.001} & 0.763 & 112.363 \\ 
Stay-ahead & \textbf{0.008} & \textbf{0.001} & 0.718 & 136.182 \\ 

\hline
\multicolumn{5}{c}{\textbf{Lab2Car + PPO} (lane follow)} \\
\hline
Baseline & 0.062 & 0.022 & 2.147 & \textbf{205.178} \\ 
Tracking & 0.051 & 0.014 & \textbf{0.617} & 187.514 \\ 
Map & 0.049 & 0.002 & 0.644 & 187.046 \\ 
Stay-behind & 0.020 & \textbf{0.001} & 1.563 & 148.713 \\ 
Stay-ahead & \textbf{0.016} & \textbf{0.001} & 1.478 & 162.141 \\ 

\hline
\multicolumn{5}{c}{\textbf{Lab2Car + IDM} (lane follow)} \\
\hline
Baseline & 0.060 & 0.031 & 2.469 & \textbf{239.961} \\ 
Tracking & 0.054 & 0.015 & \textbf{0.840} & 210.634 \\ 
Map & 0.057 & 0.002 & 0.863 & 209.957 \\ 
Stay-behind & \textbf{0.027} & \textbf{0.001} & 1.795 & 162.121 \\ 
Stay-ahead & 0.031 & 0.004 & 1.705 & 175.928 \\ 

\hline
\multicolumn{5}{c}{\textbf{Lab2Car + A*} (unstructured)} \\
\hline
Baseline       &             0.02 &             0  &      \textbf{0.98} &      \textbf{35.40} \\
Stay-behind     &             \textbf{0.01} &       0 &       \textbf{0.99} &             26.83 \\
\hline
\end{tabular}
\begin{minipage}{0.48\textwidth}
\vspace{1mm}
\footnotesize
\textbf{Coll}, front collisions. \textbf{Road}, off-road violations. \textbf{Accel}, longitudinal acceleration violations. \textbf{Dist}, total distance travelled. Values averaged across 1602 lane follow scenarios 
and 413 unstructured PUDO scenarios.
\end{minipage}
\end{table}

We first evaluated closed-loop performance on 30-s snippets of real-world drive logs generated by Motional AVs in Las Vegas (Tab.~\ref{tab:closed_loop_resim}). 
We used 1602 lane follow/ACC scenarios for Urban Driver, PPO, and IDM, and 413 unstructured PUDO scenarios for A*.
We used the Object Sim simulator from Applied Intuition \cite{simian}, which performs realistic high-fidelity physics simulation of the entire AV stack. 



As expected, more advanced configurations of Lab2Car consistently improved safety across all planners. \texttt{Stay-ahead} improved progress over \texttt{Stay-behind} while maintaining and, in some cases, improving safety. Note that collisions cannot be reduced to zero, since the constraints rely on predictions, which maybe inaccurate. 

In most cases, improvements in safety came at the expense of progress and comfort. We hypothesized that this is related to contradictory and occasionally unsatisfiable constraints, such as tracking an accelerating trajectory while braking for an obstacle. To investigate this, we evaluated \texttt{Stay-ahead} without the tracking references (Tab.~\ref{tab:no_tracking}). This resulted in improved progress with small reductions in safety, suggesting that this trade-off can be navigated by relaxing constraints.

\begin{table}[htbp]
\centering
\caption{Closed-loop simulations with/without tracking}
\label{tab:no_tracking}
\begin{tabular}{l|cccc}  
\hline
  &   \multicolumn{2}{c}{\textbf{Safety}} & \multicolumn{1}{c}{\textbf{Comfort}}  & \multicolumn{1}{c}{\textbf{Progress}}\\
\textbf{Config} & Coll (\#)  $\downarrow$ & Road (\#) $\downarrow$ & Accel (\#) $\downarrow$ & Dist (m) $\uparrow$ \\

\hline
\multicolumn{5}{c}{\textbf{Lab2Car + Urban Driver}, Stay-ahead} \\
\hline
With tracking & \textbf{0.008} & \textbf{0.001} & \textbf{0.718} & 136.182 \\ 
W/o tracking & 0.019 & \textbf{0.001} & 1.021 & \textbf{182.939} \\ 

\hline
\multicolumn{5}{c}{\textbf{Lab2Car + PPO}, Stay-ahead} \\
\hline
With tracking & \textbf{0.016} & \textbf{0.001} & \textbf{1.478} & 162.141 \\ 
W/o tracking & 0.028 & 0.002 & 2.345 & \textbf{171.433} \\ 

\hline
\multicolumn{5}{c}{\textbf{Lab2Car + IDM}, Stay-ahead} \\
\hline
With tracking & 0.031 & 0.004 & \textbf{1.705} & 175.928 \\ 
W/o tracking & \textbf{0.029} & \textbf{0.003} & 2.717 & \textbf{188.236} \\ 

\hline
\end{tabular}
\begin{minipage}{0.48\textwidth}
\vspace{1mm}
\footnotesize
Notation as in Tab~\ref{tab:closed_loop_resim}.
\end{minipage}
\vspace{-3mm}
\end{table}

\subsection{Real-world driving}
\label{sec:real}

\begin{figure*}
    \centering
    \includegraphics[width=1\textwidth,trim={4 0 25 0},clip]{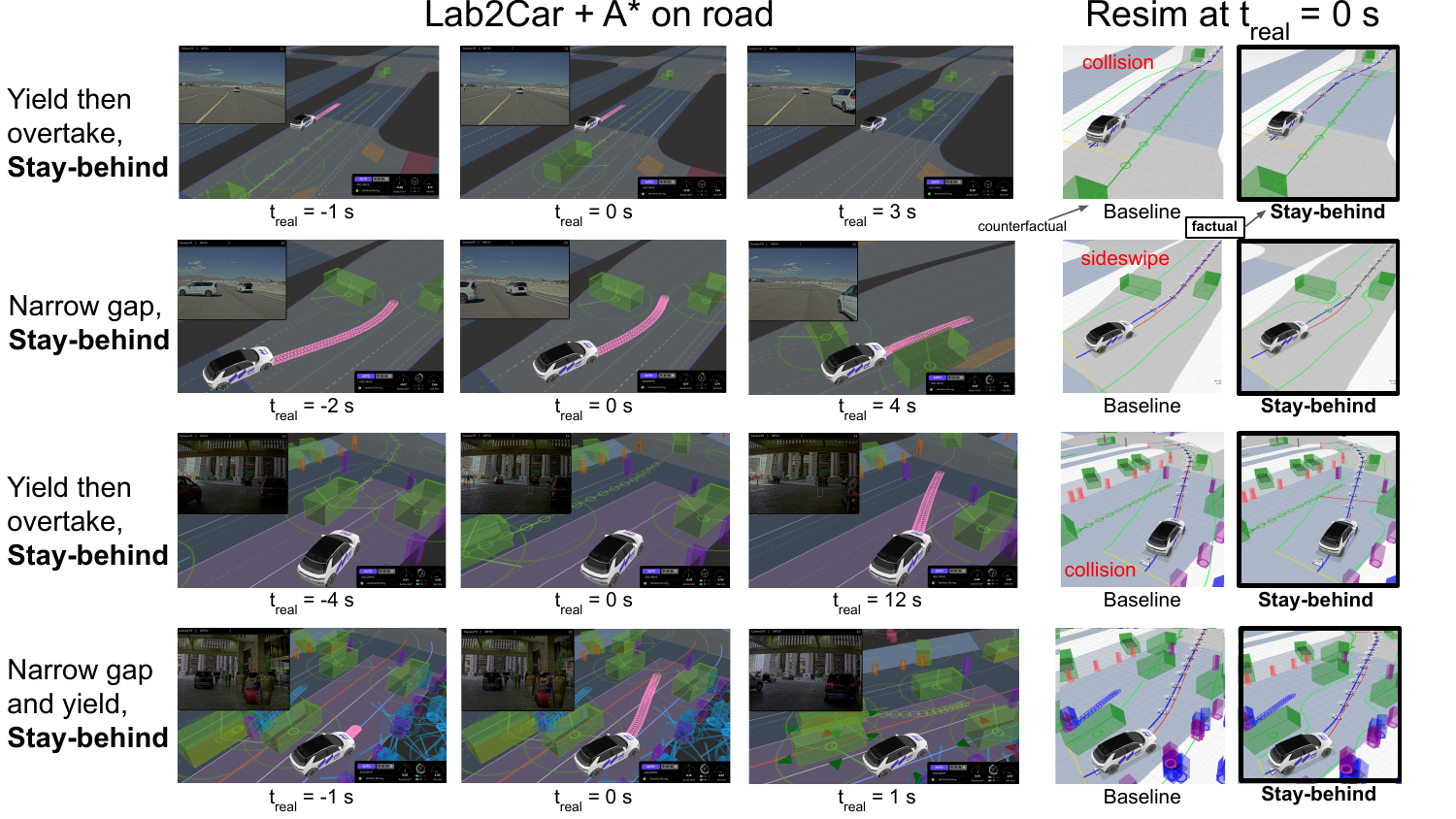}
    \caption{Lab2Car + A* on the road: private track (top two rows) and public road (bottom two rows). Notation as in Fig.~\ref{fig:urbandriveronroad}.}
    \label{fig:houdinironroad}
\vspace{-3mm}
\end{figure*}

We deployed Urban Driver and A* using Lab2Car on Motional IONIQ 5 self-driving cars in Las Vegas. Perception, prediction \cite{afshar2023pbp}, and mapping inputs were provided by the corresponding modules of the Motional AV stack. All drive tests were conducted with an experienced safety driver ready to take over in case of unsafe behavior. 
Representative scenarios are included in the supplementary video.

\begin{table}[htbp]
\centering
\caption{On-road results}
\label{tab:on_road}
\begin{tabular}{l|cccc}  
\hline
  &   \multicolumn{1}{c}{\textbf{Safety}} & \multicolumn{1}{c}{\textbf{Comfort}}  & \multicolumn{2}{c}{\textbf{Progress}} \\
\textbf{Config} & CPTO (\#) & Accel  (\#) & Dist (m) & Time (s)    \\
\hline
\multicolumn{5}{c}{\textbf{Lab2Car + Urban Driver} (private track)} \\
\hline
Map            &             2 &                    7 &          264.4 &                     75  \\ 
Stay-behind     &             0 &                    27 &         1327.0 &                    378   \\ 
Stay-ahead           &             0          &           45 &         1834.1 &                    638    \\
\hline
\multicolumn{5}{c}{\textbf{Lab2Car + A*} (private track)} \\
\hline
Stay-behind     &             0 &                    4 &           772.4 &                    567  \\
\hline
\multicolumn{5}{c}{\textbf{Lab2Car + A*} (public road)} \\
\hline
Stay-behind     &             0 &                    6 &           641.3  &                    7022   \\
\hline
\end{tabular}
\begin{minipage}{0.48\textwidth}
\vspace{1mm}
\footnotesize
Abbreviations as in Tab.~\ref{tab:closed_loop_resim}. \textbf{CPTO}, collision-preventative takeover by the safety driver. Values totaled for each configuration.
\end{minipage}
\vspace{-3mm}
\end{table}

\textbf{Lab2Car + Urban Driver on private track}: We staged 8 ACC and 9 cut-in scenarios on a private test track (Tab.~\ref{tab:on_road}, top). As a baseline, we used the \texttt{Map} configuration, which resulted in two safety-related takeovers as the planner failed to stop for the vehicle ahead (Fig.~\ref{fig:urbandriveronroad}, top row). Resimulations (resims) confirmed that the ML trajectory would have indeed resulted in a collision, which the \texttt{Stay-behind}  configuration would have prevented (Fig.~\ref{fig:urbandriveronroad}, top right). 
Accordingly, repeating the scenarios in the \texttt{Stay-behind} configuration resulted in safe stopping behind the lead vehicle (Fig.~\ref{fig:urbandriveronroad}, middle row). 
Performance was similar using the \texttt{Stay-ahead} configuration (Fig.~\ref{fig:urbandriveronroad}, bottom row). Overall, the planner slowed and/or stopped successfully for 6/8 ACC and 9/9 cut-in scenarios. 

While Lab2Car corrected the unsafe behavior of Urban Driver, the planner had comfort issues and repeatedly got stuck outputting stationary trajectories after stopping, for reasons unrelated to Lab2Car. We therefore chose not to deploy Lab2Car + Urban Driver on public roads.

\textbf{Lab2Car + A* on private track}: We similarly staged 4 overtake, 5 yield-then-overtake, and 9 narrow gap scenarios using the \texttt{Stay-behind} configuration (Tab.~\ref{tab:on_road}, middle; Fig.~\ref{fig:houdinironroad}, top two rows). The planner performed successfully in all scenarios, with the exception of getting stuck in one narrow gap scenario. There were no safety-related takeovers.

\textbf{Lab2Car + A* on public roads}:
After rigorously evaluating Lab2Car + A* in simulation and on the private test track, we deployed it on the Las Vegas strip. The experimental planner was geofenced to casino PUDO areas -- dense, unstructured environments with pedestrians, slowly moving and parked vehicles, traffic cones, and obstacles. It was deployed as part of a larger planning system that included an additional lane-based planner and a selection mechanism for arbitrating between the two. We only report results for periods when Lab2Car + A* was driving the AV.

On public roads, Lab2Car + A* showed performance similar to the private test track (Tab.~\ref{tab:on_road}, bottom; Fig.~\ref{fig:houdinironroad}, bottom two rows). It handled scenarios with multiple lane changes, jaywalking pedestrians, and multiple passing vehicles (see video). There were no safety-related takeovers.

\section{Conclusion}

In this work, we introduced Lab2Car, an optimization-based method for safely deploying experimental planners in real self-driving cars. We demonstrated the versatility of our approach by using it to deploy a ML-based planner and a classical search-based planner, neither of which could provide safety, comfort, or even kinematic feasibility on its own. Results from large-scale closed-loop simulations showed that Lab2Car can improve safety for a wide range of experimental planners. 

We envision two use cases for Lab2Car: as training wheels and as a planning component in its own right. For example, an under-trained Urban Driver can be initially deployed with the \texttt{Stay-behind} configuration. This can provide early signal for on-road issues that would be difficult to detect in simulation, such as issues with comfort or starting from stop. Unsafe behavior masked by Lab2Car can be revealed by examining discrepancies between the trajectory sketch and the final trajectory, as well as by resimulation. As Urban Driver matures, the configuration can be relaxed to \texttt{Stay-ahead}, \texttt{Map}, and finally \texttt{Tracking}, allowing Lab2Car to support the capabilities of a powerful ML planner. Alternatively, Lab2Car can become part of the final planning system. For example, our A* planner outputs a path and does not take dynamic actors into account, which means it has to be deployed with the \texttt{Stay-behind} configuration.


Lab2Car streamlines the path from incubating an idea in the lab to testing it on the car, offering early insights into the real-world performance of experimental planners at the initial stages of prototyping. This can allow researchers in academia and industry to focus on promising ideas and rule out dead ends before investing too much effort in polishing them. We believe this can dramatically accelerate progress towards resolving the planning bottleneck in autonomous driving and making a driverless future for all a reality.


\section*{Acknowledgments}

We thank Raveen Ilaalagan for on-road deployment; Mahimana Bhatt for metrics; Samuel Findler, Dmitry Yershov, Sang Uk Lee, and Caglayan Dicle for Urban Driver; Lixun Lin, Boaz Floor, Hans Andersen, and Titus Chua for the A* planner; Cedric Warny, Napat Karnchanachari, Shakiba Yaghoubi, Kate Pistunova, and Forbes Howington for feedback on the paper; and the countless other Motional scientists and engineers who contributed to this project directly or indirectly.

\bibliographystyle{IEEEtran}
\bibliography{IEEEabrv, references}

\input{supplement.tex}

\end{document}

%% file: supplement.tex
\newpage 

\supplementarytitle
\label{sec:supplement}

Here we provide additional background and details about how Lab2Car works and how it was evaluated.

\section{Related works}


\subsection{Classical planning}

Classical approaches treat the planning problem as graph search \cite{lavalle2006planning,Paden_2016} or an optimization problem \cite{Paden_2016}. Examples include A* \cite{Ziegler_2008_IVS}, RRT \cite{Lavalle_2000_ICRA,Paden_2016}, PRM* \cite{karaman2011sampling}, dynamic programming \cite{Montemerlo_2009_UC}, MPCC \cite{ferranti2019safevru}. Since the objective function is often hand-designed, parameter tuning is a painstaking process and generalization is often poor. As result, classical planners are often confined to very simple domains (e.g., lane follow \cite{treiber2000congested}).

To ensure feasible trajectories, classical planners often incorporate vehicle dynamics models in the transition function (e.g., a bicycle model \cite{mcnaughton2011motion}). Safety and comfort are often achieved via the objective function. For graph search, this is often computed locally at each transition \cite{karaman2011sampling,mcnaughton2011motion}, pruning bad states on-the-fly. Optimization-based approaches frame motion planning as a constrained optimization problem that is often solved using MPC, which can solve jointly for safety, comfort, and feasibility \cite{borrelli2005mpc,urmson2008autonomous}. Hybrid approaches combine the two techniques, e.g. using graph search to get a coarse-level path and MPC to optimize the local state transitions \cite{li2021mpc}. Control barrier functions (CBFs) have been widely used as an alternative to MPC to ensure safety efficiently \cite{ames2019control,notomista2020,seo2022}. Crucially, for classical planners, safety, comfort, and feasibility are often uniquely intertwined with the motion planning logic, resulting in bespoke planning systems that cannot be readily unbundled to benefit the deployment of less sophisticated experimental planners.
pose

\subsection{Learning-based planning}

Learning-based (or ML-based) approaches attempt to learn a driving policy from real or simulated data. Examples include imitation learning \cite{Pomerleau_1988_NIPS, Bojarski_2016_arxiv, Bansal_2018_ChauffeurNet,kendall2019learning,scheel2022urban}, reinforcement learning \cite{Dosovitskiy_2017_CoRL, Riedmiller_2007_Frontiers, Kendall_2019_ICRA, Chen_2020_CoRL, Chen_2021_ICCV}, inverse reinforcement learning \cite{Abbeel_2004_ICML,Huang_2021_ITS,phan2023driveirl}, and combinations thereof \cite{lu2023imitation}. Recently, these approaches have been boosted techniques from generative AI \cite{mao2023gpt,huang2023diffusion,seff2023motionlm,wang2024omnidrive,xu2024drivegpt4,pan2024vlp}. However, pure ML planning often cannot guarantee safety, comfort, or even basic dynamic feasibility, necessitating additional layers that often originate in classical planning.

One approach is to restrict the solution space to a discrete set of trajectories generated to satisfy the desired constraints using hand-engineered rules \cite{phan-minh_2020_cvpr}. DriveIRL \cite{phan2023driveirl} uses classification to select from a set of kinematically feasible trajectories that follow the lane and do not lead to unavoidable collisions. SafetyNet \cite{Vitelli_2022_ICRA} infers a single trajectory which is projected onto a set of feasible lane fallow trajectories. While this approach has facilitated real-world deployment of ML planners like DriveIRL and SafetyNet, it does not scale to complex scenarios, such as navigation in unstructured environments, as the trajectory space becomes intractably large.

Alternatively, ML can be used to approximate different components of classical planners, such is the value function in MCTS \cite{hoel2019combining,chekroun2024mbappe} or the transition function in MPC \cite{rokonuzzaman2021model}. More commonly, classical optimization techniques such as a PID controller \cite{chen2019,naveed2021} or MPC \cite{wabersich2021} are used post-process the output of a ML planner. Most relevant to our work, DiffStack \cite{karkus2023} and DIPP \cite{huang2023} include a differential MPC as the final layer of a neural AV stack. In addition to ensuring feasibility, safety, and comfort, this allows for the parameters of the entire AV stack it to be learned end-to-end. However, this comes at the cost of increased complexity and tight coupling between the MPC and the planning module, making the MPC less portable to other planners, especially if they are not ML-based. Furthermore, to ensure differentiability, the MPC cost function needs to be smooth, which makes it challenging to impose hard constraints, such as staying on the road or avoiding collisions. This can weaken the safety guarantees, while potentially introducing undesirable side effects, such as unnecessarily nudging the AV away from obstacles. Notably, to the best of our knowledge, neither approach has been used for real-world deployment.

\section{Maneuver extraction}
\label{sec:examples}

\begin{figure}
\centering
\fontsize{8}{10}\selectfont
    \includegraphics[width=0.5\textwidth,trim={25 195 170 0},clip]{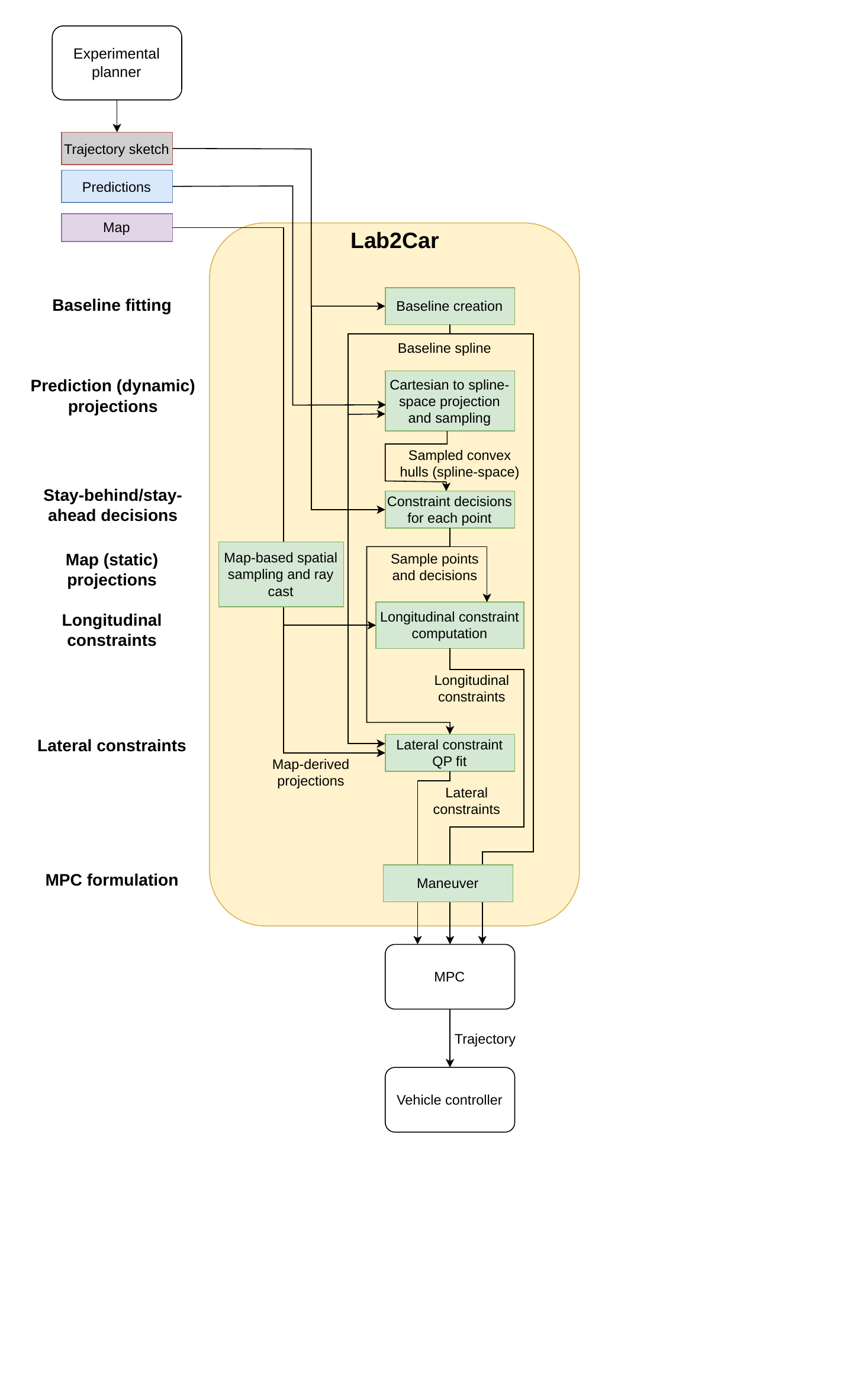}
\caption{Lab2Car maneuver extraction flow diagram.}
\label{fig:flowchart}
\end{figure}

A flowchart illustrating maneuver extraction is shown in Fig.~\ref{fig:flowchart}.

\subsection{Control point optimization}

The regularization matrix $R$ in Eq.~\ref{eq:control_point_optimization} is:

\begin{equation}
R_{N-2 \times N}=\begin{bmatrix}
-1 & 2 & -1 & 0 & \hdots \\
0 & -1 & 2 & -1 & \hdots \\
0 & 0 & -1 & 2 & \hdots \\
\vdots & \vdots & \vdots & \\
\end{bmatrix}, 
\end{equation}

\subsection{Spline-space conversion}

Cartesian points are projected onto the spline by finding the closest point. This is performed iteratively for each point by approximating the spline locally as a circle and then finding the closest point of the circle to the query point (Alg.~\ref{alg:findClosestPoint}).

\begin{algorithm}
\caption{Iterative closest point search on spline}\label{alg:findClosestPoint}
\begin{algorithmic}[1]
\small
\State \textbf{Input:} Spline $spline$, Query Point $q_p$
\State \textbf{Output:} Closest Point on Spline

\State $sampled\_points \gets \text{SamplePoints}(spline, num\_samples)$ 
\State $closest_p \gets \text{argmin}_{p} \; \text{EuclideanDistance}(sampled\_points, q_p)$ 

\For{$i = 1$ to $3$} 
    \State $center, radius \gets \text{ApproximateAsCircle}(spline, closest_p)$
    \State $delta_p \gets \text{ComputeDelta}(center, radius, q_p, spline, closest_p)$
    \State $closest_p \gets closest_p + delta_p$
\EndFor

\State \textbf{return} $spline(closest_p)$
\end{algorithmic}
\end{algorithm}

\subsection{Lateral constraint fitting}

The cost function additionally incorporates a regularization matrix \(R\), which governs the rate of change of the spline. This regularization is represented by:

\begin{equation}
R=\begin{bmatrix}
-1 & 1 & 0 & 0 & \hdots \\
0 & -1 & 1 & 0 & \hdots \\
0 & 0 & -1 & 1 & \hdots \\
\vdots & \vdots & \vdots & \\
\end{bmatrix}
\end{equation}

This matrix serves to smooth transitions and maintain a consistent gradient across the spline, aiding in the stabilization of the fitted trajectory. This approach ensures that the spline tubes accurately reflect the spatial constraints imposed by the surrounding environment.


\section{MPC formulation}

The MPC non-linear constraint function is given by:

\begin{equation}
\mathbf{C(x)}=\begin{bmatrix}
0.5 \cdot \mathbf{W} + \mathbf{FL} \cdot \sin(\omega) - \spatialfun{\text{left}}{p + \mathbf{FL} \cdot \cos(\omega)} \\
0.5 \cdot \mathbf{W} + \mathbf{FL} \cdot \sin(\omega) + \spatialfun{\text{right}}{p + \mathbf{FL} \cdot \cos(\omega)} \\
0.5 \cdot \mathbf{W} - \mathbf{RL} \cdot \sin(\omega) - \spatialfun{\text{left}}{p + \mathbf{RL} \cdot \cos(\omega)} \\
0.5 \cdot \mathbf{W} - \mathbf{RL} \cdot \sin(\omega) + \spatialfun{\text{right}}{p + \mathbf{RL} \cdot \cos(\omega)} \\
\vdots
\end{bmatrix}
\end{equation}

\section{Experiments}

\subsection{Experimental planners}
\label{sec:planners}

We evaluated Lab2Car together with four motion planners that cover two main types of approaches used in the literature and industry: ML and classical. 

\subsubsection{Urban Driver}
\label{sec:urbandriver}

We used Urban Driver \cite{scheel2022urban}, a ML-based motion planner which uses regression to learn to imitate an expert driver using closed-loop rollouts. We adapted the open-source version of Urban Driver made public by the authors and trained it on the nuPlan dataset \cite{caesar2021nuplan,karnchanachari2024towards} [\url{https://github.com/motional/nuplan-devkit/}]. We restricted the training set to lane follow and ACC scenarios. The input to the model was a vectorized map and a vectorized representation of the other agents in the scene (vehicles, pedestrians, bicyclists, etc.) relative to the AV. The output was a sequence of 16 $(x,y)$-waypoints corresponding to an 8-s open-loop AV trajectory at 2 Hz. We trained the model with multistep training: for each scenario, we rolled out the model for 13 s at 2 Hz and supervised the output open-loop trajectory at each time step with the corresponding ground truth expert trajectory using an L2 loss.

\subsubsection{Proximal policy optimization (PPO)}
\label{sec:ppo}

PPO is a reinforcement learning method for training an agent to maximize a reward function based on simulated experience. We trained an open-source PPO implementation from Stable Baselines 3 \cite{stable-baselines3} on lane follow and ACC scenarios using a custom simulator. Specifically, we procedurally generated scenarios with heuristically driven agents, accumulating roughly 1 billion samples, corresponding to about one year of simulated driving. The state was the longitudinal position, velocity, and acceleration of the AV and the lead vehicle (if any). The action was the longitudinal jerk control. The transition integrated jerk along the lane centerline based on the AV state. The reward function included a comfort term (penalizing jerk and acceleration), a collision term, and a desired clearance term (to the lead vehicle). During inference, we rolled out the model for 8 s at 2 Hz, using predictions to get the lead vehicle at each future time step.

\subsubsection{Intelligent driver model (IDM)}
\label{sec:idm}

The IDM \cite{treiber2000congested} is a classical planner for lane following and ACC. It computes longitudinal acceleration in closed form as a function of the longitudinal position and speed of the AV and the lead vehicle (if any). Similarly to PPO, during inference, we rolled out the model for 8 s at 2 Hz, using predictions to get the lead vehicle at each future time step.

\subsubsection{A* search}
\label{sec:astar}

We used A* \cite{hart1968formal}, a classical graph search algorithm which we adapted for freespace motion planning. Vertices were defined by a discrete space of $(x, y, \theta)$-poses and edges were defined by feasible transitions between poses, with weights representing the relative cost of motion. The heuristic was an (approximate) lower bound of the cost-to-go to the target. The input to the planner was a starting pose, a target pose, and an occupancy grid with the drivable area and the static agents. The output was a sequence of $(x, y, \theta)$-waypoints connecting the starting pose to the target pose. The waypoints did not have corresponding timestamps, so unlike Urban Driver, the output corresponded to a purely spatial path with no temporal dimension. This planner was designed for freespace planning in unstructured environments (e.g., parking lots) and ignored dynamic agents, instead relying on Lab2Car for active collision avoidance.

\subsection{Perception, Prediction, and Mapping}

For simulated scenarios, object-oriented representations of the scene from the simulator were fed to the motion planner and to Motional's in-house prediction module \cite{afshar2023pbp}, which generated unimodal 8-s predictions for each actor in the scene at each iteration of the simulation. These predictions were used by Lab2Car to compute lateral and longitudinal constraints for each open-loop time step of the maneuver. For real-world driving, we similarly used Motional's in-house perception system to recreate an object-oriented representation of the scene from camera and lidar sensor data. Perception ran at 20 Hz, while prediction ran at 10 Hz, at the same rate as planning. We also assume access to a high-definition map. While robust perception, prediction, and mapping are critical for both motion planners and for Lab2Car, they are also integral components of a modular AV stack and are widely used in industry. 
Therefore we omit their details as they fall beyond the scope of this paper.

\subsection{Scenarios}
\label{sec:scenarios}

We evaluated Lab2Car using the Object Sim simulator (formerly Simian) from Applied Intuition \cite{simian} on 2000+ closed-loop resim scenarios based on real-world drive logs generated by Motional AVs in Las Vegas. Each scenario consisted of a 30-s snippet from the drive log downsampled to 10 Hz. The first 3 s of each scenario consisted of a ``warm-up'' period during which the system was run in open loop but its inputs and outputs were clamped to (i.e., replayed from) the log. For the remaining 27 s, the system was run in closed loop at 10 Hz. The open-loop trajectory from each iteration was tracked by a VDS controller, which was designed to mimic the dynamics of the real AV. All other actors were replayed from the log, i.e. they were nonreactive to any deviations of AV behavior resulting from the closed-loop simulation.

Given the complementary strengths of the two motion planners, we evaluated them on different sets of scenarios:

\subsubsection{Nominal Scenarios}

We evaluated Lab2Car + Urban Driver, PPO, and IDM on 1602 nominal lane follow and ACC scenarios on the Las Vegas Strip similar to those from the nuPlan training set. They were meant to illustrate how Lab2Car can act as a rule-based safety wrapper and smoother for a well-trained but imperfect ML planner.

\subsubsection{Freespace Scenarios}

We evaluated Lab2Car + A* on 413 scenarios from pick-up/drop-off (PuDo) areas of multiple casinos off the Las Vegas Strip. The unstructured PuDo environments included irregularly parked cars, slow-moving traffic, pedestrians jaywalking and getting on and off of vehicles. They were meant to demonstrate how Lab2Car can be used in conjunction with a purely spatial planner to allow progress while ensuring safety in unstructured environments.

\subsection{Metrics}
\label{sec:metrics}

To illustrate how Lab2Car impacts safety, comfort, and progress, we calculated the following metrics:

\begin{itemize}
    \item Front collisions (\textbf{Coll}): instances when the AV collided with a vehicle in front of it.
\footnote{We exclude rear collisions from reporting as they would be misleading, producing many false positives due to the nonreactive behavior of the other actors in our simulations.}
    \item Off-road (\textbf{Road}): instances when any point of the AV footprint was outside the driveable area.
    \item Acceleration violation (\textbf{Accel}): instances when the (absolute) longitudinal acceleration was above a certain threshold. When decelerating, the threshold was 2.5 m/s$^2$ for speeds $<$ 10 m/s and 1.5 m/s$^2$ for speeds $>$ 20 m/s, with linear interpolation for in-between speeds.  When accelerating, the threshold was 2.0 m/s$^2$ for speeds $<$ 10 m/s and 1.0 m/s$^2$ for speeds $>$ 15 m/s, with linear interpolation for in-between speeds.
    \item Distance traveled (\textbf{Dist}): total distance traveled by the AV for the entire duration of the scenario.
        \item Collision-preventative takeovers (\textbf{CPTO}): 
    instances when the safety driver took over to prevent a collision (on-road evaluation only). 
\end{itemize}

For simulations, metrics were calculated separately for each scenario based on the closed-loop trajectory of the AV and the replayed trajectories of the other actors along the entire duration of the scenario. We aggregated each metric across scenarios by averaging for each configuration.

For real-world driving, metrics were computed separately for each run with a given planner and Lab2Car configuration.


\subsection{Ablations}
\label{sec:ablations}

To study the relative importance of different Lab2Car components, we evaluated different ablated versions (configurations) of Lab2Car (Fig.~\ref{fig:ablations}). We present them here in order from simplest to most complex, with each subsequent configuration enabling a superset of the components enabled in the previous configuration:



\begin{itemize}
    \item \texttt{Baseline}: We only fitted the baseline spline to the initial trajectory sketch. No lateral or longitudinal constraints were imposed based on the map or other actors in the scene. No tracking costs were imposed and the AV was allowed to accelerate to the speed limit. In this configuration, the system simply generates dynamically feasible trajectories that follow the spatial (but not the temporal) intent of the initial trajectory sketch. 
    \item \texttt{Tracking}: In addition to fitting the baseline (as in the \texttt{Baseline} configuration), longitudinal tracking costs were imposed at waypoint projections onto the baseline. These encouraged the AV to reach the corresponding speed derived from the initial directory sketch at each corresponding waypoint. No lateral or longitudinal constraints were imposed. In this configuration, the system tracks the intent of the initial trajectory sketch both spatially and temporally.
    \item \texttt{Map}: In addition to fitting the baseline and tracking the waypoints (as in the \texttt{Tracking} configuration), constraints were imposed based on the map. Specifically, the lateral tube was restricted to the boundaries of the drivable area. If the width of the lateral tube shrunk below a certain threshold (2 m), indicating that the AV will surely violate the lateral bounds and drive off the road, a longitudinal constraint was imposed. In this configuration, the system tracks the intent of the initial trajectory sketch, while remaining on the drivable area.
    \item \texttt{Stay-behind}: In addition to fitting the baseline, tracking the waypoints, and staying on the drivable area (as in the \texttt{Map} configuration), constraints were imposed based on the other actors in the scene. Specifically, lateral constraints were imposed on actors within 4 m of the baseline, to ensure the AV circumvents them. If an actor was too close to the baseline ($<$ 2 m), indicating that driving around them is unsafe or infeasible, a stay-behind longitudinal constraint was imposed. In this configuration, the AV was constrained to always stay behind other actors crossing its path and pass after they have moved out of the way.
    \item \texttt{Stay-ahead}: This included all components from the \texttt{Stay-behind} configuration, namely fitting the baseline, tracking the waypoints, staying on the drivable area, and constraints based on the other actors. However, the longitudinal constraints were computed based on where the AV was expected to be along the baseline, according to the waypoint projections. Specifically, if the AV was expected to be behind an actor at time $t$, a \textit{stay-behind} longitudinal constraint was imposed to ensure that AV indeed remains behind at time $t$, as in the \texttt{Stay-behind} configuration. However, if the AV was expected to be in front of the actor at time $t$, a \textit{stay-ahead} longitudinal constraint was imposed instead, ensuring that AV passes the actor by time $t$. Thus the AV could either \textit{stay behind} or \textit{stay ahead} other actors crossing its path, allowing for a diverse set of behaviors, including lane merging and cut-ins.
\end{itemize}

In stimulation, we evaluated Urban Driver in all configurations. For A*, we only used the \texttt{Baseline} and \texttt{Stay-behind} configurations. Since our A* implementation produces a path rather than a trajectory, there are no timestamps associated with the waypoints, so tracking references and stay-ahead constraints could not be computed. Additionally, since the occupancy grid \textit{a priori} constrains the search to only consider poses on the driveable area, the \texttt{Map} configuration produced identical performance to the \texttt{Baseline} configuration.

On the road, we evaluated Urban Driver in the \texttt{Map}, \texttt{Stay-behind}, and \texttt{Stay-ahead} configurations, given the relatively poor performance of the \texttt{Baseline} and \texttt{Tracking} configurations in simulation.